\pdfoutput=1
\documentclass{article}

\usepackage[final, nonatbib]{neurips_2020}

\usepackage[utf8]{inputenc} % allow utf-8 input
\usepackage[T1]{fontenc}    % use 8-bit T1 fonts
\usepackage{hyperref}       % hyperlinks
\usepackage{url}            % simple URL typesetting
\usepackage{booktabs}       % professional-quality tables
\usepackage{amsfonts}       % blackboard math symbols
\usepackage{nicefrac}       % compact symbols for 1/2, etc.
\usepackage{microtype}      % microtypography
\usepackage{amsmath}        % amsmath
\usepackage{multirow}       % multirow
\usepackage[frak=euler]{mathalfa}       % Math fonts
\usepackage[numbers, square, sort]{natbib}
\usepackage{tikz}  % tikz for drawing
\usepackage{tikz-network}

\title{Investigating Estimated Kolmogorov Complexity as a Means of Regularization for Link Prediction}

\author{%
  Paris D. L. Flood, Ramon Vi\~{n}as, Pietro Li\`{o}\\
  Department of Computer Science and Technology\\
  University of Cambridge\\
  \texttt{\{pdlf3,rv340,pl219\}@cam.ac.uk}
}

\begin{document}

\maketitle

\begin{abstract}
Link prediction in graphs is an important task in the fields of network science and machine learning. We investigate a flexible means of regularization for link prediction based on an approximation of the Kolmogorov complexity of graphs that is differentiable and compatible with recent advances in link prediction algorithms. Informally, the Kolmogorov complexity of an object is the length of the shortest computer program that produces the object. Complex networks are often generated, in part, by simple mechanisms; for example, many citation networks and social networks are approximately scale-free and can be explained by preferential attachment. A preference for predicting graphs with simpler generating mechanisms motivates our choice of Kolmogorov complexity as a regularization term. In our experiments the regularization method shows good performance on many diverse real-world networks, however we determine that this is likely due to an aggregation method rather than any actual estimation of Kolmogorov complexity.
\end{abstract}

\section{Introduction}
Network models have become an indispensable tool to study complex systems of discrete objects and their interactions. Network science has been applied with great success to a variety of scientific disciplines, often resulting in rich data sets that can be studied under the auspices of machine learning. A significant area of research that has emerged from the study of networks is link prediction between nodes based on incomplete instances of data \citep{surveyMartinez2016}. The importance of link prediction techniques is underscored by its breadth of applications to topics such as protein function anticipation \citep{holme2005role}, friendship identification for social network users \citep{6413904} and scientific collaboration inference \citep{liben2007link}.

Many complex networks have simple causal mechanisms that underlie their generation. For example, it has been theorized that scale-free networks such as the World Wide Web are often generated by a system of preferential attachment whereby new nodes are more likely to attach to nodes that are already well connected \citep{Barabasi509}. The small-world property, which states that any two nodes can reach each other via a short path, is present in many real-world phenomenon such as social networks and can be artificially generated by the simple Watts-Strogatz model \citep{watts1998collective}. The observation that complex networks can be characterized by such simple mechanisms raises the possibility of incorporating complexity biases for network modelling.

In recent years, graph neural networks (GNNs) have proven to be highly adept at solving link prediction problems \citep{wu2020comprehensive,xu2018powerful,kipf2016variational}. By leveraging relational inductive biases to obtain high-level node representations, architectures such as graph auto-encoders can decode meaningful, unseen links from their latent representations \citep{wu2020comprehensive, kipf2016variational}. Another benefit of GNNs is that they can be trained in the canonical way by defining a loss function and employing gradient-based learning methods. This flexible training framework affords an opportunity to incorporate information about the causal generating mechanisms of networks through a regularization term. 

In this work, we investigate penalizing graphs with large estimated Kolmogorov complexity to encourage the creation of graphs with simpler generating mechanisms. We accomplish this by proposing a differentiable regularization term for link prediction based on a popular method to estimate the Kolmogorov complexity of graphs \cite{zenil-decomposition, zenil2015two}. The Kolmogorov complexity of an object is the length of the shortest computer program that outputs that object and is uncomputable, hence the need for an estimation method. Kolmogorov complexity and the related field of algorithmic probability rigorously define a notion of simplicity, and have been used to discover simple neural networks with a marked ability to generalize \citep{SCHMIDHUBER1997857}. By proposing Kolmogorov complexity as a regularizer on graph outputs, we aim to augment the ability of models such as graph neural networks to generalize when predicting links. In our experiments we learn that although our proposed regularization term shows good results, it very likely works for reasons related to an aggregation routine more similar to classical entropy rather than any direct estimation of the Kolmogorov complexity.

\paragraph{Related Work} In \cite{ALGprobNonDiff19}, Kolmogorov complexity was proposed by Hern\'{a}ndez-Orozco et al. as a regularization term with a weighting parameter in conjunction with a general loss function. Similar to Schmidhuber's approach in \cite{SCHMIDHUBER1997857}, Hern\'{a}ndez-Orozco et al.'s form of regularization pushes a model towards low complexity, thus increasing the algorithmic probability of the model. In our methodology, we attempt to use Kolmogorov complexity to regulate the output of the model, rather than the model. We take this approach because we would like to reward the model for learning to generate objects from simpler rule sets. Hern\'{a}ndez-Orozco et al. also point out that a Kolmogorov penalty function, as they have defined it, cannot be optimized by gradient-based methods \cite{ALGprobNonDiff19}. We overcome this problem by using a probabilistic interpretation of the regularization term. 

In order to approximate the gradient of our estimated Kolmogorov complexity based regularization term, we use a method based on perturbing the predicted adjacency matrix. The study of the algorithmic causality of an object by means of a perturbation calculus on the object's estimated Kolmogorov complexity was pioneered by Zenil et al. in \cite{ZENIL2019sciencecd}. Their approach was recently used to define a highly effective unsupervised algorithm for identifying generating mechanisms in graphs \citep{zenil2019causaldeconv}.

\paragraph{Paper Outline} This paper is structured as follows. Section \ref{sec:back_inf} briefly presents several fundamental concepts in algorithmic information theory and discusses a method for approximating Kolmogorov complexity that we later use in our methodology. Section \ref{sec:methodology} describes our estimated Kolmogorov complexity regularization term for link prediction that is fully compatible with standard differentiable approaches to training neural networks, such as backpropagation. Section \ref{sec:experiment} presents experiments on five diverse real-world networks and Section \ref{sec:discussionconclusion} discusses the experimental results (and why the effectiveness is likely not due to any actual estimation of Kolmogorov complexity) and concludes the paper by considering challenges and future work.

\section{Background Information}
\label{sec:back_inf}
Algorithmic information theory (AIT) primarily studies the irreducible information content of objects and was charmingly summarized by one of its founders, Gregory Chaitin, as ``the result of putting Shannon's information theory and Turing's computability theory into a cocktail shaker and shaking vigorously \citep{auckquote}". The information content or complexity of an object is measured by the size, in bits, of the shortest computer program that can compute the object. For example, consider the following two binary strings: \texttt{10101010101010101010101010101010} and \texttt{00011011011111010101000100110110}. The first string has a simple pattern which can be described concisely as: \texttt{print `10' 16 times}. The second string does not have a clear pattern and likely has no simpler description than merely printing the string itself. In more formal terms, the Kolmogorov complexity of a string $s$ is $K(s) = \min\{|p|: U(p) = s \} $ where $p$ is a program of length $|p|$ bits that, when run on a universal Turing machine $U$, outputs $s.$ Kolmogorov complexity is invariant to the choice of $U$ up to an additive constant independent of the choice of $s.$ 

Closely related to Kolmogorov complexity is the concept of algorithmic probability, a method for assigning a universal prior probability to objects. Consider a program $p$ that produces a binary string $s$ when run on a universal prefix-free Turing machine $U.$ The universal prior probability for each string $s$ is defined as:
\[
m(s) = \sum_{p: U(p) = s}2^{-|p|}
\]
As $T$ is a universal prefix-free Turing machine, the group of valid programs on $U$ are a prefix-free set and thus, by Kraft's inequality, the sum is bounded by one. Kolmogorov complexity and algorithmic probability are beautifully linked through Levin's Coding Theorem which gives the following result:
\[
-\log_2 m(s) = K(s) + \mathcal{O}(1)
\]
Algorithmic probability is, in part, guided by Epicurus' principle of multiple explanations (if several theories are consistent with the data, retain them all) and Occam's razor (among theories consistent with the data, choose the simplest) \citep{hutter2004universal}. The prior probability $m(s)$ satisfies these principles by assigning a non-zero probability to every string and giving a higher probability to strings with shorter generating programs. For more information on the field of algorithmic information theory in general, we refer the reader to the following references \citep{li2008introduction, hutter2004universal, kolmogorov1968three, solomonoff1964formal, solomonoff1964formal2, chaitin1966length, chaitin1969length}.

\subsection{Approximating Kolmogorov Complexity}
Both Kolmogorov complexity and algorithmic probability are uncomputable for reasons related to the halting problem, therefore approximations are required. Statistical lossless compression algorithms are a popular approach to estimate Kolmogorov complexity \citep{zenil2020review}. Lossless compression techniques like the Lempel-Ziv-Welch (LZW) algorithm clearly provide intuitive estimates of the Kolmogorov complexity of an object, but suffer from an inability to capture meaning beyond classical Shannon information theory \citep{zenil-decomposition,zenil2020review}. 

\paragraph{Coding Theorem Method} The Coding Theorem Method (CTM) provides a straightforward approximation to the Kolmogorov complexity of an object that captures algorithmic features rather than merely statistical features \citep{soler-toscano}. The CTM directly approximates the algorithmic probability of small strings by exploring the large space of Turing machines with a fixed number of symbols and states. Let $(n,2)$ be the class of all $n\text{-state}$ $2\text{-symbol}$ Turing machines $T$ using the Turing machine formalism outlined in the busy beaver game \citep{rado1962non}. The CTM defines the following function for a binary string $s$:
\[
D_{(n,2)}(s) = \frac{|\{T \in (n,2) : T\; \text{produces}\; s\}|}{|\{T \in (n,2) : T \; \text{halts}\}|}
\]
Of course, in general it is impossible to know if a machine will halt; however, for the $2$ symbol case the largest number of steps taken before halting are known up to $n = 4$ (and theorized for $n=5$) \citep{soler-toscano}. Therefore, $D_{(n,2)}(s)$ can be computed for small $n$ using brute-force. Using an approximate form of Levin's Coding Theorem, the CTM estimate of Kolmogorov complexity, denoted by $CTM_{(n,2)}(s)$, is given as:
\[
CTM_{(n,2)}(s) = -\log_2D_{(n,2)}(s)
\]

\paragraph{Block Decomposition Method} Unfortunately, it is ultimately uncomputable to use the CTM approximation on large objects due to the rapid growth of the busy beaver function. To address this limitation, the Block Decomposition Method (BDM) \citep{zenil-decomposition} extends the CTM via an aggregation rule designed to reconstruct the Kolmogorov complexity of a large object from its smaller components. The BDM has been applied with great success to problems in machine learning and causality \citep{zenil2019causaldeconv, ALGprobNonDiff19} and is described as follows. For a given binary string $s$, decompose the string into the multiset $\mathcal{S}_s = \{ s_1, s_2, \dots, s_{\frac{|s|}{r}}\}$ where $s_i$ are consecutive slices from $s$ of size $r.$ The value of $r$ is chosen to be small enough so that the CTM can compute an approximation to the Kolmogorov complexity of the slice (we assume the length of $s$ is divisible by $r$). Let $\mathcal{U}_s$ be the set of unique values in $\mathcal{S}_s$ and let $c_u$ be the number of times slice $u \in \mathcal{U}_s$ appears in $\mathcal{S}_s.$ The BDM is based on the following intuition: if the CTM approximates the Kolmogorov complexity for each slice $u$, then a program with an estimated complexity of $\sum_{u \in \mathcal{U}_s} CTM_{(n,2)}(u)$ can be used to generate all of the unique building blocks of $s.$ The number of times each slice $u$ appears in $s$ can the be specified in $\log_2(c_u)$ bits, thus the BDM approximation for the Kolmogorov complexity of $s$ is:
\[
BDM_{(n,2)}(s) = \sum_{u \in \mathcal{U}_s} CTM_{(n,2)}(u) + \log_2(c_u)
\]

\paragraph{Approximating Graph Complexity} The BDM can be extended to approximate the Kolmogorov complexity of a graph by applying a two-dimensional variant of the BDM to the graph's binary adjacency matrix $\mathbf{A}$ \citep{zenil-decomposition, zenil2015two}. The CTM component of the BDM approximation is computed using Turing machines that run on a 2-dimensional tape and produce arrays rather than strings. The BDM approximation of the graph's Kolmogorov complexity $BDM_{(n,2)}(\mathbf{A})$ is computed over a partition of $\mathbf{A}$ into block matrices small enough to have a CTM value (for a more explicit formulation please see Section \ref{prac_diff_form}). Using the adjacency matrix as the descriptor of a graph introduces a potential challenge as adjacency matrices corresponding to isomorphic graphs can have different Kolmogorov complexity. However, this discrepancy is bounded by a constant independent of the choice of graph \citep{mitzenil15} and in practice the BDM works very well as a Kolmogorov complexity estimator for networks \cite{zenil2019causaldeconv, morzy2017measuring, zenil2014correlation}.

Larger values of $n$ allow for CTM estimates of larger arrays, and in turn lead to better BDM approximations \citep{zenil-decomposition}. Therefore, for the remainder of this paper we will replace the notation $BDM_{(n,2)}$ with $K_{BDM}$ where $n$ is assumed to be the largest number of states for which there are CTM values available. We will also exclusively use the function $K_{BDM}$ in reference to the two-dimensional variant of the BDM.

\section{Methodology}
\label{sec:methodology}

\paragraph{Notation} We consider an unweighted graph $\mathcal{G} = (\mathcal{V},\mathcal{E})$ with $N = |\mathcal{V}|$ nodes. The $N \times N$ adjacency matrix $\mathbf{A}$ of $\mathcal{G}$ has elements $a_{ij} \in \{0,1\}.$ Given a learning algorithm $\mathcal{M}$ that predicts links (in the form of an adjacency matrix), we denote the output of $\mathcal{M}$ as an $N \times N$ matrix $\mathbf{\tilde{A}}.$ The elements of $\mathbf{\tilde{A}}$ have been mapped to the open interval $(0,1)$ by the output activation function of $\mathcal{M}.$ We will treat $\mathbf{\tilde{A}}$ as a matrix of Bernoulli parameters where $\tilde{a}_{ij}$ represents the independent probability that there is an edge from node $i$ to node $j.$   The reasoning for treating $\mathbf{\tilde{A}}$ as a matrix of probabilities will be discussed in Section \ref{prac_diff_form}. When referring to the Bernoulli random variable parameterized by $\tilde{a}_{ij}$ we will write $\tilde{\mathfrak{a}}_{ij}.$ Additionally, when referring to the matrix of independent Bernoulli random variables parameterized by the values in $\mathbf{\tilde{A}}$, we will write $\mathbfrak{\tilde{A}}.$

\subsection{Regularized Loss Function}
Let $\mathcal{L}$ denote a general loss function used to train $\mathcal{M}$ over $\mathbf{A}_{Train}$, a noisy or restricted view of $\mathbf{A}$. For example, a reasonable choice of $\mathcal{L}$ is the binary cross entropy loss function with weighting to account for a sparsity of edges.

\paragraph{Kolmogorov Regularization} Given a learning algorithm that predicts links in a graph, we define \textit{Kolmogorov-regularized} functions as the class of loss functions with the form:
\begin{equation}
    \hat{\mathcal{L}} = \mathcal{L} + \lambda \cdot \mathbb{E}[K(\mathbfrak{\tilde{A}})]
\end{equation}
where $\lambda \in \mathbb{R}^{+}$ is a weighting hyperparameter and $\mathbb{E}[K(\mathbfrak{\tilde{A}})]$ is the expected Kolmogorov complexity of $\mathbfrak{\tilde{A}}$. Because the Kolmogorov complexity of an object is uncomputable, we rely on the BDM to produce an approximation $\mathbb{E}[K_{BDM}(\mathbfrak{\tilde{A}})]$ of the expected Kolmogorov complexity of $\mathbfrak{\tilde{A}}$. We denote the class of loss functions that use this approximation to Kolmogorov regularization as:
\begin{equation}
    \hat{\mathcal{L}}_{BDM} = \mathcal{L} + \lambda \cdot \mathbb{E}[K_{BDM}(\mathbfrak{\tilde{A}})]
\end{equation}

\subsection{Practical Differentiable Formulation}
\label{prac_diff_form}
Let us partition the binary adjacency matrix $\mathbf{A}$ into blocks of size $R \times R$ as follows:
\[ 
\mathbf{A} = 
\begin{bmatrix}
  \mathbf{A}_{11} & \mathbf{A}_{12} & \cdots & \mathbf{A}_{1N'} \\
  \mathbf{A}_{21} & \mathbf{A}_{22} & \cdots & \mathbf{A}_{2N'} \\
  \vdots          & \vdots          & \ddots & \vdots          \\
  \mathbf{A}_{N'1} & \mathbf{A}_{N'2} & \cdots & \mathbf{A}_{N'N'}
\end{bmatrix}
\]
where $N' = \nicefrac{N}{R}.$ We have made the mild assumption that $N$ is divisible by $R$; if this is not the case we pad $\mathbf{A}$ with zeros (see Section \ref{sec:ExpDes}).  We refer to the multiset of all blocks in $\mathbf{A}$ as $\mathcal{A}_\mathbf{A} = \{\mathbf{A}_{11}, \mathbf{A}_{21}, \mathbf{A}_{12}, \dots, \mathbf{A}_{N'N'}\}$ and the set of unique elements in $\mathcal{A}_\mathbf{A}$ as $\mathcal{U}_{\mathbf{A}}$. Recall that the BDM approximation of the Kolmogorov complexity of a binary adjacency matrix $\mathbf{A}$ is:
\[
K_{BDM}(\mathbf{A}) = \sum_{\mathbf{U} \in \mathcal{U}_{\mathbf{A}}} CTM \left( \mathbf{U} \right) + \log_2 \left( c_\mathbf{U} \right)
\]
where $c_\mathbf{U}$ is the number of times a block $\mathbf{U} \in \mathcal{U}_{\mathbf{A}}$ appears in $\mathcal{A}_\mathbf{A}.$ Because we would like our regularization term to be used with gradient-based training algorithms such as backpropagation, we must alter the BDM approximation to be differentiable. This requirement motivates our designation of the model output $\mathbf{\tilde{A}}$ as an adjacency matrix of edge probabilities. By treating $\mathbfrak{\tilde{A}}$ as a collection of $N^2$ independent Bernoulli random variables paramterized by $\mathbf{\tilde{A}}$, we have a regularization term $\mathbb{E}[K_{BDM}(\mathbfrak{\tilde{A}})]$ that is clearly differentiable with respect to the elements of $\mathbf{\tilde{A}}$. However, this decision also introduces a computational complexity problem. To appreciate this problem, note that each of the ${N'}^2$ blocks in $\mathbfrak{\tilde{A}}$ has a unique probability mass function over the $2^{R^2}$ possible binary matrices of size $R \times R.$ Therefore, there are ${\left(2^{R^2} \right) }^{{N'}^2} = 2^{N^2}$ unique probabilities to be computed in order to directly determine $\mathbb{E}[K_{BDM}(\mathbfrak{\tilde{A}})]$ (this is also apparent from the fact that there are $2^{N^2}$ possible realizations of $\mathbfrak{\tilde{A}}$).

\paragraph{Monte Carlo Perturbation}
To mitigate the computational complexity problem, we adopt a Monte Carlo approach where we sample $m$ times from $\mathbfrak{\tilde{A}}.$ However, instead of approximating $\mathbb{E}[K_{BDM}(\mathbfrak{\tilde{A}})]$ we use the samples to directly approximate the gradient $\nabla \mathbb{E}[K_{BDM}(\mathbfrak{\tilde{A}})].$ Consider the partial derivative of $\mathbb{E}[K_{BDM}(\mathbfrak{\tilde{A}})]$ with respect to $\tilde{a}_{ij}:$
\begin{equation*}
\begin{split}
\frac{\partial \mathbb{E}[K_{BDM}(\mathbfrak{\tilde{A}})]}{\partial \tilde{a}_{ij}} & = \frac{\partial}{\partial\tilde{a}_{ij}} \cdot \mathbb{P} \left( \tilde{\mathfrak{a}}_{ij} = 1 \right) \cdot \mathbb{E}[K_{BDM}(\mathbfrak{\tilde{A}}) | \tilde{\mathfrak{a}}_{ij} = 1]\\
& + \frac{\partial}{\partial\tilde{a}_{ij}} \cdot\mathbb{P} \left( \tilde{\mathfrak{a}}_{ij} = 0 \right) \cdot \mathbb{E}[K_{BDM}(\mathbfrak{\tilde{A}}) | \tilde{\mathfrak{a}}_{ij} = 0] \\
& = \frac{\partial}{\partial\tilde{a}_{ij}} \cdot  \tilde{a}_{ij} \cdot \mathbb{E}[K_{BDM}(\mathbfrak{\tilde{A}}) | \tilde{\mathfrak{a}}_{ij} = 1]\\
& + \frac{\partial}{\partial\tilde{a}_{ij}} \cdot \left( 1 - \tilde{a}_{ij} \right) \cdot \mathbb{E}[K_{BDM}(\mathbfrak{\tilde{A}}) | \tilde{\mathfrak{a}}_{ij} = 0] \\
& = \mathbb{E}[K_{BDM}(\mathbfrak{\tilde{A}}) | \tilde{\mathfrak{a}}_{ij} = 1] - \mathbb{E}[K_{BDM}(\mathbfrak{\tilde{A}}) | \tilde{\mathfrak{a}}_{ij} = 0]
\end{split}
\end{equation*}

Let $\mathbf{\tilde{A}}^{(1)}, \mathbf{\tilde{A}}^{(2)}, \dots, \mathbf{\tilde{A}}^{(m)}$ denote $m$ binary matrices sampled from $\mathbfrak{\tilde{A}}.$ For each sample $\mathbf{\tilde{A}}^{(k)}$ we can partition the matrix into $R \times R$ blocks and compute a frequency table of all the different blocks in $\mathcal{O}(N^2)$ time. We will use the notations $\mathbf{\tilde{A}}_{ij = 1}^{(k)}$ and $\mathbf{\tilde{A}}_{ij = 0}^{(k)}$ to denote a sample $\mathbf{\tilde{A}}^{(k)}$ that has element ${\tilde{a}}_{ij}^{(k)}$ set to either $1$ or $0$, regardless of the original value of ${\tilde{a}}_{ij}^{(k)}$. We also have access to a pre-computed lookup table of CTM values for every possible $R \times R$ binary matrix. Note that $R << N$ and is generally set at a fixed value of $4$ (see Section \ref{sec:ExpDes}); therefore, in our analysis it will be treated as a constant. Using both the lookup table and the frequency table, the difference $K_{BDM}(\mathbf{\tilde{A}}_{ij = 1}^{(k)}) - K_{BDM}(\mathbf{\tilde{A}}_{ij = 0}^{(k)})$ can be computed in $\mathcal{O}(1)$ time. This is accomplished by using the binary string of the values in the $R \times R$ block matrix containing ${\tilde{a}}_{ij}^{(k)}$ as the index key for both tables, and then simply incrementing and decrementing BDM values based on the existing frequencies. The average value of $K_{BDM}(\mathbf{\tilde{A}}_{ij = 1}^{(k)}) - K_{BDM}(\mathbf{\tilde{A}}_{ij = 0}^{(k)})$ approaches $\frac{\partial \mathbb{E}[K_{BDM}(\mathbfrak{\tilde{A}})]}{\partial \tilde{a}_{ij}}$ as more samples are taken. Therefore, if we compute the value of $K_{BDM}(\mathbf{\tilde{A}}_{ij = 1}^{(k)}) - K_{BDM}(\mathbf{\tilde{A}}_{ij = 0}^{(k)})$ for each element ${\tilde{a}}_{ij}^{(k)}$ we have effectively sampled from the gradient $\nabla \mathbb{E}[K_{BDM}(\mathbfrak{\tilde{A}})]$ in $\mathcal{O}(N^2)$ time.

\paragraph{Loss Function Incorporation} Let  ${\nabla \mathbb{E}[K_{BDM}(\mathbfrak{\tilde{A}})]}^{(k)}$ denote a sample from $\nabla \mathbb{E}[K_{BDM}(\mathbfrak{\tilde{A}})]$ and for simplicity of notation we write the sample mean of the gradient as: 
\[
\mathbf{\bar{G}}_{\mathbf{\tilde{A}}, m} = \frac{1}{m} \sum_{k = 1}^m {\nabla \mathbb{E}[K_{BDM}(\mathbfrak{\tilde{A}})]}^{(k)}
\]
In order to incorporate the sample mean of the gradient into the gradient of the loss function we simply multiply each element in $\mathbf{\tilde{A}}$ with its corresponding element in $\mathbf{\bar{G}}_{\mathbf{\tilde{A}}, m}$ and the weighting parameter $\lambda$, then sum these products back into the loss function. Note that despite the notation, the elements of $\mathbf{\bar{G}}_{\mathbf{\tilde{A}}, m}$ are treated as constants. Our loss function is summarized below as:

\begin{equation}
\label{Eq: central_K_reg_method}
     \hat{\mathcal{L}}_{BDM}^{*} = \mathcal{L} + \lambda \cdot  \mathbf{1}^{T} \left( \mathbf{\tilde{A}} \odot \mathbf{\bar{G}}_{\mathbf{\tilde{A}}, m} \right) \mathbf{1}
\end{equation}
where $\odot$ denotes element-wise multiplication and $\mathbf{1}$ is the column vector of $N$ ones.

\section{Experiments}
\label{sec:experiment}
In order to assess the performance of our method, we measure the impact of the regularization term $\lambda \cdot  \mathbf{1}^{T} \left( \mathbf{\tilde{A}} \odot \mathbf{\bar{G}}_{\mathbf{\tilde{A}}, m} \right) \mathbf{1}$ on the ability of standard GNN frameworks to predict links. More specifically, we test the regularization term on a graph auto-encoder (GAE) and a variational graph auto-encoder (VGAE) \citep{kipf2016variational}. For both models, we follow the designs used in \citep{kipf2016variational} where the encoders are two-layer graph convolutional networks (GCN) \citep{kipf2016semi} with 32 and 16 hidden units, respectively. The decoders produce the edge probabilities by computing the sigmoid of the inner product of the latent node embeddings. As for the loss functions, the GAE network is trained on the binary cross-entropy loss with weighting proportional to the ratio of negative to positive labels. The VGAE network uses the same loss as a reconstruction term, but includes an additional Kullback-Leibler divergence term to measure the discrepancy between the approximation of the posterior and the latent prior, which we define as an isotropic Gaussian distribution with unit variance. We also test a limited version of the regularization term where the weights from the pre-computed CTM lookup table are replaced by a single constant value (we use the average value of the entire CTM table for reasons discussed Section \ref{sec:ExpDes}). This alternate regularization term, which we will refer to as constant weight (CW) regularization, serves as a control to elucidate whether improvements in performance stem from the CTM or the aggregation rule in the BDM.

\subsection{Link Prediction on Real-World Networks}
We perform our experimentation on five different real-world networks: a network of links between Wikipedia pages on chameleons \citep{rozemberczki2019multiscale}, a road transportation network from Chicago \citep{konect:2017:tntp-ChicagoRegional, konect:tntp-chicago1, konect:tntp-chicago2}, the Cora citation network of scientific publications \citep{ yang2016revisiting}, a protein-protein interaction network from PDZBase \citep{beuming2005pdzbase, konect:2017:maayan-pdzbase}, and a network of co-purchases of US political books \citep{politicalbooks}. These data sets were chosen to represent a broad range of applications with highly different generating mechanisms. As both the GAE and VGAE models require a node feature matrix, we use an appropriately sized identity matrix as a dummy input for each of the five networks. Additionally, each network is processed to be undirected and contain no self-loops.
%
%\begin{table}
%  \caption{Overview of the networks used in our experiments.}
%  \label{Ta: network-table}
%  \centering
%  \begin{tabular}{lccc}
%    \toprule
%    \textbf{Network} & \textbf{Node Count}    & \textbf{Edge Count}    & \textbf{Category} \\
%    \midrule
%    Chameleon    & $2277$    & $31421$ & Internet\\
%    Chicago    & $1467$    & $1298$ & Transportation\\
%    Cora    & $2708$    & $5429$ & Citation\\
%    PDZBase    & $212$    & $244$ & Protein Interaction\\
%    Political Books    & $105$    & $441$ & Purchasing\\
%    \bottomrule
%  \end{tabular}
%\end{table}
%

\paragraph{Experiment Design}
\label{sec:ExpDes} Our experiment design is largely based on that of \citep{kipf2016variational, kipf2020deep}. We begin by dividing each of the five networks into training, validation, and testing data sets. The training input is the original adjacency matrix with $80\%$ of the edges randomly retained. Because the graph convolutional operator requires the adjacency matrix to be updated with self-loops along the diagonal, our training label is simply the training input summed with the identity matrix. The validation set consists of half of the $20\%$ of original edges not selected for training along with an equal number of false edges that do not exist in the original graph. All of the true edges and false edges are randomly selected. The test set consists of the remaining original edges along with an equal number of random false edges that do not exist in either the original graph or the set of false validation edges. Note that different random splits will, of course, give slightly different results.

We randomly initialize both the GAE and the VGAE models using Glorot initialization \citep{glorot2010understanding} and perform multiple trials to account for different initializations. We employ two standard metrics for binary classification: area under the ROC curve (AUC) and average precision (AP). After splitting the data sets, we establish preliminary results for both models without any regularization over $10$ trials on each of the five validation sets. All trials described in this paper are run for $1000$ epochs. During each trial, we save the model weights for both the maximum validation AUC and AP scores.

Using these preliminary results we search for a $\lambda$ for the Kolmogorov regularization term on the validation sets in a simple manner, with care taken to make sure this process is not overly tedious. The starting point for the search is the inverse of the square of the node count as the BDM can potentially grow quadratically with the node count (until the CTM dictionary is exhausted). We then proceed to search in proportional increments until neither increasing nor decreasing $\lambda$ leads to a significant increase in validation performance for Kolmogorov regularization. Table \ref{Ta: lambda-val} contains the values of $\lambda$ in both models for each of the five networks. Throughout our experiments we use $m = 1$ sample to approximate the gradient $\nabla \mathbb{E}[K_{BDM}(\mathbfrak{\tilde{A}})].$ Increasing the value of $m$ does not seem to improve results significantly, but does slow down training as the regularization term takes $\mathcal{O}(mN^2)$ time to compute. $R$ is set to $4,$ the largest value for which there are binary CTM array estimates available \citep{pybdm}. If $N$ is not divisible by $R$ we can simply pad $\mathbf{A}$ with zeros; this has a negligible effect on the BDM as $R = 4 << N$.

\begin{table}
  \caption{Summary of $\lambda$ values used for each data set.}
  \label{Ta: lambda-val}
  \centering
  \begin{tabular}{rccccc}
    \toprule
    \textbf{Network}        & Chameleon & Chicago & Cora    & PDZBase & Political Books \\
    \midrule
    \textbf{$\lambda$ Value}  & $5 \times 10^{-7}$      & $1 \times 10^{-5}$    & $4 \times 10^{-7}$    & $1 \times 10^{-4}$    & $3 \times 10^{-5}$ \\
    \bottomrule
  \end{tabular}
\end{table}

After the $\lambda$ values have been determined, we train all five data sets using Kolmogorov regularization. For both models, we repeat this process for $10$ trials per data set, saving the model weights that yield the maximum validation AUC and AP scores for each trial and data set. This process is also repeated for the constant weight regularization with the exception that the same $\lambda$ values are used from the Kolmogorov regularization (there is no search on the validation sets). Of course, this means that the constant weight regularization values could be higher (if they had their own tailored $\lambda$ values). However, because we chose the constant weight to be equal to the average value of the CTM table, we get good enough results on the constant weight regularization term to fufill its role as a control (see Section \ref{sec:discussionconclusion}). Finally, we run all the saved validation model weights (without regularization, with Kolmogorov regularization, and with constant weight regularization) on the corresponding network test sets. We report the means and standard errors on the test sets for both AUC and AP scores in Tables \ref{Ta: summary-auc-gae}, \ref{Ta: summary-auc-vgae}, \ref{Ta: summary-ap-gae}, and \ref{Ta: summary-ap-vgae}. Bold values indicate that highest range of that value according to the given precision of standard error is at least as good as the lowest range of the other two values in the row.

\begin{table}
  \caption{Link prediction results for AUC metric on the GAE architecture.}
  \label{Ta: summary-auc-gae}
  \centering
 \begin{tabular}{lccc}
    \toprule
    \multirow{2}{*}{\textbf{Network}} &
      \multicolumn{3}{c}{\textbf{GAE}}\\
      & {No Reg.} & {Kol. Reg.} & {CW Reg.} \\
      \midrule
    Chameleon & $98.22 \pm 0.01$ & $\mathbf{98.90 \pm 0.02}$ &  $\mathbf{98.91 \pm 0.02}$\\
    Chicago & $76.38 \pm 1.21$ & $\mathbf{88.00 \pm 0.13}$ &   $\mathbf{88.23 \pm 0.13}$\\
    Cora & $81.85 \pm 0.48$ & $\mathbf{82.60 \pm 0.24}$ &   $\mathbf{83.06 \pm 0.30}$\\
    PDZBase & $71.51 \pm 2.48$ & $\mathbf{83.14 \pm 0.44}$ &   $\mathbf{83.77 \pm 0.56}$\\
    Political Books & $83.70 \pm 0.41$ & $\mathbf{88.94 \pm 0.43}$ &   ${87.95 \pm 0.49}$\\
    \bottomrule
\end{tabular}
\end{table}

\begin{table}
  \caption{Link prediction results for AUC metric on the VGAE architecture.}
  \label{Ta: summary-auc-vgae}
  \centering
 \begin{tabular}{lccc}
    \toprule
    \multirow{2}{*}{\textbf{Network}} &
      \multicolumn{3}{c}{\textbf{VGAE}} \\
      & {No Reg.} & {Kol. Reg.} & {CW Reg.}\\
      \midrule
    Chameleon & $98.17 \pm 0.02$ & $\mathbf{98.82 \pm 0.02}$  & $\mathbf{98.85 \pm 0.02}$\\
    Chicago & $82.40 \pm 0.79$ & $\mathbf{88.11 \pm 0.08}$ & ${87.82 \pm 0.19}$\\
    Cora & ${82.56 \pm 0.28}$ & $\mathbf{82.94 \pm 0.30}$ & $\mathbf{83.54 \pm 0.34}$\\
    PDZBase & $75.28 \pm 1.64$ & $\mathbf{83.89 \pm 0.63}$ & $\mathbf{84.79 \pm 0.43}$\\
    Political Books & $86.61 \pm 0.43$ & $\mathbf{88.96 \pm 0.33}$ & $\mathbf{88.50 \pm 0.39}$\\
    \bottomrule
\end{tabular}
\end{table}

\begin{table}
  \caption{Link prediction results for AP metric on the GAE architecture.}
  \label{Ta: summary-ap-gae}
  \centering
 \begin{tabular}{lccc}
    \toprule
    \multirow{2}{*}{\textbf{Network}} &
      \multicolumn{3}{c}{\textbf{GAE}}\\
      & {No Reg.} & {Kol. Reg.} & {CW Reg.} \\
      \midrule
    Chameleon & $98.48 \pm 0.02$ & $\mathbf{98.96 \pm 0.02}$ & $\mathbf{98.97 \pm 0.01}$ \\
    Chicago & $78.35 \pm 1.03$ & $\mathbf{86.17 \pm 0.31}$ & $\mathbf{86.18 \pm 0.21}$ \\
    Cora & ${86.09 \pm 0.23}$ & ${86.21 \pm 0.19}$ & $\mathbf{86.65 \pm 0.14}$  \\
    PDZBase & $\mathbf{75.04 \pm 1.42}$ & $\mathbf{73.80 \pm 2.25}$ & $\mathbf{77.92 \pm 2.28}$  \\
    Political Books & $80.75 \pm 0.45$ & ${89.28 \pm 0.40}$ & $\mathbf{90.67 \pm 0.32}$  \\
    \bottomrule
\end{tabular}
\end{table}

\begin{table}
  \caption{Link prediction results for AP metric on the VGAE architecture.}
  \label{Ta: summary-ap-vgae}
  \centering
 \begin{tabular}{lccc}
    \toprule
    \multirow{2}{*}{\textbf{Network}} &
      \multicolumn{3}{c}{\textbf{VGAE}} \\
      & {No Reg.} & {Kol. Reg.} & {CW Reg.}\\
      \midrule
    Chameleon  &  $98.52 \pm 0.02$ & $\mathbf{98.88 \pm 0.02}$ & $\mathbf{98.91 \pm 0.01}$\\
    Chicago &  $82.59 \pm 1.10$ & $\mathbf{86.22 \pm 0.15}$ & ${85.94 \pm 0.10}$\\
    Cora &  $86.26 \pm 0.25$ & $\mathbf{86.76 \pm 0.21}$ & $\mathbf{86.79 \pm 0.25}$\\
    PDZBase &  $\mathbf{76.02 \pm 1.20}$ & $\mathbf{77.16 \pm 2.52}$ & $\mathbf{73.33 \pm 2.13}$\\
    Political Books &  $85.04 \pm 0.33$ & $\mathbf{91.08 \pm 0.25}$ & ${90.54 \pm 0.21}$\\
    \bottomrule
\end{tabular}
\end{table}

Training was performed on an Intel i7-4790k CPU with 8GB of RAM and an Nvidia GTX 970 GPU. Computation of the gradient sample for the regularization term was done entirely on the CPU, although this process should scale well on a GPU. Each trial of $1000$ epochs took from approximately $30$ seconds (Political Books) to approximately $35$ minutes (Cora) depending on the size of the network.

\section{Discussion and Conclusion}
\label{sec:discussionconclusion}
In Tables \ref{Ta: summary-auc-gae}, \ref{Ta: summary-auc-vgae}, \ref{Ta: summary-ap-gae}, and \ref{Ta: summary-ap-vgae}, both the Kolmogorov and constant weight regularization terms appear to be effective for link prediction tasks on a broad variety of data sets. In particular, the performance gains on the Chameleon, Chicago and Political Books networks were impressive (relative to the standard errors) when the graph neural networks were trained with regularization. Note that despite already being very high, the results on the Chameleon data set were significantly superior with regularization as the standard error ranges for this network were very small. The Cora data set displayed some marginal increases, but in comparison to other networks the improvements were not as drastic. Regularization did have a notable positive effect on the AUC score of the PDZBase network, but the AP score did not improve significantly. The volatility shown on the PDZBase network was likely due to its small edge count.

However, it seems unlikely that any of the gains from the Kolmogorov regularization were due to the CTM as the constant weight control regularization performed essentially just as well, even without tailored $\lambda$ values which would have likely further improved its performance. Because the CTM is a direct attempt to estimate the Kolmogorov complexity, it seems safe to conclude that the Kolmogorov regularization term does not improve results because of an appeal to Kolmogorov complexity, but rather a more simple, frequency based regularity (similar to classical entropy). We can offer a few possible explanations for the lack of contribution from the CTM, and thus estimation of Kolmogorov complexity, to the improvements from regularization. It is possible that the CTM cannot overcome the constant term in Levin's Coding Theorem - a concern highlighted in a recent article by Vit{\'a}nyi \cite{vitanyi2020incomputable}. Another potential explanation is a lack of sensitivity to the differences in estimated Kolmogorov complexity for the CTM. A linear increase in bit size leads to an exponential increase in possible programs, however the regularization term still punishes this increase in a linear sense.

If the latter problem is the true issue, it could be addressed in future work by redesigning the regularization term to be more sensitive to different CTM weights (which range from about $22$ to $36$ for $R=4$). However, if the problem is actually that the CTM cannot be reliably used to estimate Kolmogorov complexity due to difficulties with the size of the constant term in Levin's Coding Theorem, then an ambitious task for future research is the development of alternative approximation methods of Kolmogorov complexity. Naturally, better approximations would seemingly allow for an increase in the quality of regularization bias towards networks with simpler generating mechanisms. 

\bibliography{bib}

\begin{thebibliography}{42}
\providecommand{\natexlab}[1]{#1}
\providecommand{\url}[1]{\texttt{#1}}
\expandafter\ifx\csname urlstyle\endcsname\relax
  \providecommand{\doi}[1]{doi: #1}\else
  \providecommand{\doi}{doi: \begingroup \urlstyle{rm}\Url}\fi

\bibitem[Mart\'{\i}nez et~al.(2016)Mart\'{\i}nez, Berzal, and
  Cubero]{surveyMartinez2016}
V\'{\i}ctor Mart\'{\i}nez, Fernando Berzal, and Juan-Carlos Cubero.
\newblock A survey of link prediction in complex networks.
\newblock \emph{ACM Comput. Surv.}, 49\penalty0 (4), December 2016.
\newblock ISSN 0360-0300.
\newblock \doi{10.1145/3012704}.
\newblock URL \url{https://doi.org/10.1145/3012704}.

\bibitem[Holme and Huss(2005)]{holme2005role}
Petter Holme and Mikael Huss.
\newblock Role-similarity based functional prediction in networked systems:
  application to the yeast proteome.
\newblock \emph{Journal of the Royal Society Interface}, 2\penalty0
  (4):\penalty0 327--333, 2005.

\bibitem[{Dong} et~al.(2012){Dong}, {Tang}, {Wu}, {Tian}, {Chawla}, {Rao}, and
  {Cao}]{6413904}
Y.~{Dong}, J.~{Tang}, S.~{Wu}, J.~{Tian}, N.~V. {Chawla}, J.~{Rao}, and
  H.~{Cao}.
\newblock Link prediction and recommendation across heterogeneous social
  networks.
\newblock pages 181--190, 2012.

\bibitem[Liben-Nowell and Kleinberg(2007)]{liben2007link}
David Liben-Nowell and Jon Kleinberg.
\newblock The link-prediction problem for social networks.
\newblock \emph{Journal of the American society for information science and
  technology}, 58\penalty0 (7):\penalty0 1019--1031, 2007.

\bibitem[Barab{\'a}si and Albert(1999)]{Barabasi509}
Albert-L{\'a}szl{\'o} Barab{\'a}si and R{\'e}ka Albert.
\newblock Emergence of scaling in random networks.
\newblock \emph{Science}, 286\penalty0 (5439):\penalty0 509--512, 1999.
\newblock ISSN 0036-8075.
\newblock \doi{10.1126/science.286.5439.509}.
\newblock URL \url{https://science.sciencemag.org/content/286/5439/509}.

\bibitem[Watts and Strogatz(1998)]{watts1998collective}
Duncan~J Watts and Steven~H Strogatz.
\newblock Collective dynamics of ‘small-world’networks.
\newblock \emph{nature}, 393\penalty0 (6684):\penalty0 440, 1998.

\bibitem[Wu et~al.(2020)Wu, Pan, Chen, Long, Zhang, and
  Philip]{wu2020comprehensive}
Zonghan Wu, Shirui Pan, Fengwen Chen, Guodong Long, Chengqi Zhang, and S~Yu
  Philip.
\newblock A comprehensive survey on graph neural networks.
\newblock \emph{IEEE Transactions on Neural Networks and Learning Systems},
  2020.

\bibitem[Xu et~al.(2018)Xu, Hu, Leskovec, and Jegelka]{xu2018powerful}
Keyulu Xu, Weihua Hu, Jure Leskovec, and Stefanie Jegelka.
\newblock How powerful are graph neural networks?
\newblock \emph{arXiv preprint arXiv:1810.00826}, 2018.

\bibitem[Kipf and Welling(2016{\natexlab{a}})]{kipf2016variational}
Thomas~N. Kipf and Max Welling.
\newblock Variational graph auto-encoders, 2016{\natexlab{a}}.

\bibitem[Zenil et~al.(2018)Zenil, Hernández-Orozco, Kiani, Soler-Toscano,
  Rueda-Toicen, and Tegnér]{zenil-decomposition}
Hector Zenil, Santiago Hernández-Orozco, Narsis~A. Kiani, Fernando
  Soler-Toscano, Antonio Rueda-Toicen, and Jesper Tegnér.
\newblock A decomposition method for global evaluation of shannon entropy and
  local estimations of algorithmic complexity.
\newblock \emph{Entropy}, 20\penalty0 (8), 2018.
\newblock ISSN 1099-4300.
\newblock \doi{10.3390/e20080605}.
\newblock URL \url{https://www.mdpi.com/1099-4300/20/8/605}.

\bibitem[Zenil et~al.(2015)Zenil, Soler-Toscano, Delahaye, and
  Gauvrit]{zenil2015two}
Hector Zenil, Fernando Soler-Toscano, Jean-Paul Delahaye, and Nicolas Gauvrit.
\newblock Two-dimensional kolmogorov complexity and an empirical validation of
  the coding theorem method by compressibility.
\newblock \emph{PeerJ Computer Science}, 1:\penalty0 e23, 2015.

\bibitem[Schmidhuber(1997)]{SCHMIDHUBER1997857}
Jürgen Schmidhuber.
\newblock Discovering neural nets with low kolmogorov complexity and high
  generalization capability.
\newblock \emph{Neural Networks}, 10\penalty0 (5):\penalty0 857 -- 873, 1997.
\newblock ISSN 0893-6080.
\newblock \doi{https://doi.org/10.1016/S0893-6080(96)00127-X}.
\newblock URL
  \url{http://www.sciencedirect.com/science/article/pii/S089360809600127X}.

\bibitem[Hern{\'a}ndez-Orozco et~al.(2019)Hern{\'a}ndez-Orozco, Zenil, Riedel,
  Uccello, Kiani, and Tegn{\'e}r]{ALGprobNonDiff19}
Santiago Hern{\'a}ndez-Orozco, Hector Zenil, J{\"u}rgen Riedel, Adam Uccello,
  Narsis~A Kiani, and Jesper Tegn{\'e}r.
\newblock Algorithmic probability-guided supervised machine learning on
  non-differentiable spaces.
\newblock \emph{arXiv preprint arXiv:1910.02758}, 2019.

\bibitem[Zenil et~al.(2019{\natexlab{a}})Zenil, Kiani, Marabita, Deng, Elias,
  Schmidt, Ball, and Tegnér]{ZENIL2019sciencecd}
Hector Zenil, Narsis~A. Kiani, Francesco Marabita, Yue Deng, Szabolcs Elias,
  Angelika Schmidt, Gordon Ball, and Jesper Tegnér.
\newblock An algorithmic information calculus for causal discovery and
  reprogramming systems.
\newblock \emph{iScience}, 19:\penalty0 1160 -- 1172, 2019{\natexlab{a}}.
\newblock ISSN 2589-0042.
\newblock \doi{https://doi.org/10.1016/j.isci.2019.07.043}.
\newblock URL
  \url{http://www.sciencedirect.com/science/article/pii/S2589004219302706}.

\bibitem[Zenil et~al.(2019{\natexlab{b}})Zenil, Kiani, Zea, and
  Tegn{\'e}r]{zenil2019causaldeconv}
Hector Zenil, Narsis~A Kiani, Allan~A Zea, and Jesper Tegn{\'e}r.
\newblock Causal deconvolution by algorithmic generative models.
\newblock \emph{Nature Machine Intelligence}, 1\penalty0 (1):\penalty0 58--66,
  2019{\natexlab{b}}.

\bibitem[Chaitin(2010)]{auckquote}
Gregory Chaitin.
\newblock Algorithmic information theory.
\newblock
  \url{https://www.cs.auckland.ac.nz/research/groups/CDMTCS/docs/ait.php?fbclid=IwAR1zYoWmVB-mdToZSlkv8wQqVQFsoCFG9rglhKVIL7uNk6iPtRtF8j6PYFo},
  2010.
\newblock Accessed: 2020-06-01.

\bibitem[Hutter(2004)]{hutter2004universal}
Marcus Hutter.
\newblock \emph{Universal artificial intelligence: Sequential decisions based
  on algorithmic probability}.
\newblock Springer Science \& Business Media, 2004.

\bibitem[Li et~al.(2008)Li, Vit{\'a}nyi, et~al.]{li2008introduction}
Ming Li, Paul Vit{\'a}nyi, et~al.
\newblock \emph{An introduction to Kolmogorov complexity and its applications},
  volume~3.
\newblock Springer, 2008.

\bibitem[Kolmogorov(1968)]{kolmogorov1968three}
Andrei~Nikolaevich Kolmogorov.
\newblock Three approaches to the quantitative definition of information.
\newblock \emph{International journal of computer mathematics}, 2\penalty0
  (1-4):\penalty0 157--168, 1968.

\bibitem[Solomonoff(1964{\natexlab{a}})]{solomonoff1964formal}
Ray~J Solomonoff.
\newblock A formal theory of inductive inference. part i.
\newblock \emph{Information and control}, 7\penalty0 (1):\penalty0 1--22,
  1964{\natexlab{a}}.

\bibitem[Solomonoff(1964{\natexlab{b}})]{solomonoff1964formal2}
Ray~J Solomonoff.
\newblock A formal theory of inductive inference. part ii.
\newblock \emph{Information and control}, 7\penalty0 (2):\penalty0 224--254,
  1964{\natexlab{b}}.

\bibitem[Chaitin(1966)]{chaitin1966length}
Gregory~J Chaitin.
\newblock On the length of programs for computing finite binary sequences.
\newblock \emph{Journal of the ACM (JACM)}, 13\penalty0 (4):\penalty0 547--569,
  1966.

\bibitem[Chaitin(1969)]{chaitin1969length}
Gregory~J Chaitin.
\newblock On the length of programs for computing finite binary sequences:
  statistical considerations.
\newblock \emph{Journal of the ACM (JACM)}, 16\penalty0 (1):\penalty0 145--159,
  1969.

\bibitem[Zenil(2020)]{zenil2020review}
Hector Zenil.
\newblock A review of methods for estimating algorithmic complexity: Options,
  challenges, and new directions, 2020.

\bibitem[Soler-Toscano et~al.(2014)Soler-Toscano, Zenil, Delahaye, and
  Gauvrit]{soler-toscano}
Fernando Soler-Toscano, Hector Zenil, Jean-Paul Delahaye, and Nicolas Gauvrit.
\newblock Calculating kolmogorov complexity from the output frequency
  distributions of small turing machines.
\newblock \emph{PLOS ONE}, 9\penalty0 (5):\penalty0 1--18, 05 2014.
\newblock \doi{10.1371/journal.pone.0096223}.
\newblock URL \url{https://doi.org/10.1371/journal.pone.0096223}.

\bibitem[Rado(1962)]{rado1962non}
Tibor Rado.
\newblock On non-computable functions.
\newblock \emph{Bell System Technical Journal}, 41\penalty0 (3):\penalty0
  877--884, 1962.

\bibitem[Zenil et~al.(2016)Zenil, Kiani, and Tegnér]{mitzenil15}
Hector Zenil, Narsis~A. Kiani, and Jesper Tegnér.
\newblock Methods of information theory and algorithmic complexity for network
  biology.
\newblock \emph{Seminars in Cell and Developmental Biology}, 51:\penalty0 32 --
  43, 2016.
\newblock ISSN 1084-9521.
\newblock \doi{https://doi.org/10.1016/j.semcdb.2016.01.011}.
\newblock URL
  \url{http://www.sciencedirect.com/science/article/pii/S1084952116300118}.
\newblock Information Theory in Systems Biology Xenopus as a model system for
  vertebrate development.

\bibitem[Morzy et~al.(2017)Morzy, Kajdanowicz, and
  Kazienko]{morzy2017measuring}
Miko{\l}aj Morzy, Tomasz Kajdanowicz, and Przemys{\l}aw Kazienko.
\newblock On measuring the complexity of networks: Kolmogorov complexity versus
  entropy.
\newblock \emph{Complexity}, 2017, 2017.

\bibitem[Zenil et~al.(2014)Zenil, Soler-Toscano, Dingle, and
  Louis]{zenil2014correlation}
Hector Zenil, Fernando Soler-Toscano, Kamaludin Dingle, and Ard~A Louis.
\newblock Correlation of automorphism group size and topological properties
  with program-size complexity evaluations of graphs and complex networks.
\newblock \emph{Physica A: Statistical Mechanics and its Applications},
  404:\penalty0 341--358, 2014.

\bibitem[Kipf and Welling(2016{\natexlab{b}})]{kipf2016semi}
Thomas~N Kipf and Max Welling.
\newblock Semi-supervised classification with graph convolutional networks.
\newblock \emph{arXiv preprint arXiv:1609.02907}, 2016{\natexlab{b}}.

\bibitem[Rozemberczki et~al.(2019)Rozemberczki, Allen, and
  Sarkar]{rozemberczki2019multiscale}
Benedek Rozemberczki, Carl Allen, and Rik Sarkar.
\newblock Multi-scale attributed node embedding, 2019.
\newblock URL
  \url{https://github.com/benedekrozemberczki/datasets#wikipedia-article-networks}.

\bibitem[kon(2017{\natexlab{a}})]{konect:2017:tntp-ChicagoRegional}
Chicago network dataset -- {KONECT}, April 2017{\natexlab{a}}.
\newblock URL \url{http://konect.uni-koblenz.de/networks/tntp-ChicagoRegional}.

\bibitem[Eash et~al.(1983)Eash, Chon, Lee, and Boyce]{konect:tntp-chicago1}
R.~W. Eash, K.~S. Chon, Y.~J. Lee, and D.~E. Boyce.
\newblock Equilibrium traffic assignment on an aggregated highway network for
  sketch planning.
\newblock \emph{Transportation Research Record}, 994:\penalty0 30--37, 1983.

\bibitem[Boyce et~al.(1985)Boyce, Chon, Ferris, Lee, Lin, and
  Eash]{konect:tntp-chicago2}
D.~E. Boyce, K.~S. Chon, M.~E. Ferris, Y.~J. Lee, K-T. Lin, and R.~W. Eash.
\newblock Implementation and evaluation of combined models of urban travel and
  location on a sketch planning network.
\newblock \emph{Chicago Area Transportation Study}, pages xii + 169, 1985.

\bibitem[Yang et~al.(2016)Yang, Cohen, and Salakhutdinov]{yang2016revisiting}
Zhilin Yang, William~W Cohen, and Ruslan Salakhutdinov.
\newblock Revisiting semi-supervised learning with graph embeddings.
\newblock \emph{arXiv preprint arXiv:1603.08861}, 2016.
\newblock URL \url{https://github.com/kimiyoung/planetoid/raw/master/data}.

\bibitem[Beuming et~al.(2005)Beuming, Skrabanek, Niv, Mukherjee, and
  Weinstein]{beuming2005pdzbase}
Thijs Beuming, Lucy Skrabanek, Masha~Y Niv, Piali Mukherjee, and Harel
  Weinstein.
\newblock Pdzbase: a protein--protein interaction database for pdz-domains.
\newblock \emph{Bioinformatics}, 21\penalty0 (6):\penalty0 827--828, 2005.

\bibitem[kon(2017{\natexlab{b}})]{konect:2017:maayan-pdzbase}
Pdzbase network dataset -- {KONECT}, April 2017{\natexlab{b}}.
\newblock URL \url{http://konect.uni-koblenz.de/networks/maayan-pdzbase}.

\bibitem[Krebs(2004)]{politicalbooks}
V.~Krebs.
\newblock Political books network data.
\newblock \url{http://www-personal.umich.edu/~mejn/netdata/}, 2004.
\newblock Accessed: 2020-06-01.

\bibitem[Kipf et~al.(2020)]{kipf2020deep}
TN~Kipf et~al.
\newblock Deep learning with graph-structured representations.
\newblock 2020.

\bibitem[Glorot and Bengio(2010)]{glorot2010understanding}
Xavier Glorot and Yoshua Bengio.
\newblock Understanding the difficulty of training deep feedforward neural
  networks.
\newblock pages 249--256, 2010.

\bibitem[Talaga(2019)]{pybdm}
Szymon Talaga.
\newblock Pybdm: Python interface to the block decomposition method.
\newblock \url{https://github.com/sztal/pybdm}, 2019.
\newblock Accessed: 2020-06-02.

\bibitem[Vit{\'a}nyi(2020)]{vitanyi2020incomputable}
Paul Vit{\'a}nyi.
\newblock How incomputable is kolmogorov complexity?
\newblock \emph{Entropy}, 22\penalty0 (4):\penalty0 408, 2020.

\end{thebibliography}

\end{document}